\documentclass[10pt,twocolumn,letterpaper]{article}

\usepackage{cvpr}
\usepackage{times}
\usepackage{epsfig}
\usepackage{graphicx}
\usepackage{subfigure}
\usepackage{amsmath}
\usepackage{tabularx}
\usepackage{amssymb}
\usepackage{mathrsfs}
\usepackage{algorithm}
\usepackage{algorithmic}
\usepackage{multirow}
\usepackage{bm}

\usepackage[pagebackref=true,breaklinks=true,letterpaper=true,colorlinks,bookmarks=false]{hyperref}
\cvprfinalcopy 

\def\cvprPaperID{2} 

\ifcvprfinal\fi
\begin{document}

\title{Self-Tuned Deep Super Resolution}

\author{Zhangyang Wang\dag, Yingzhen Yang\dag, Zhaowen Wang\ddag, Shiyu Chang\dag, \\
Wei Han\dag, Jianchao Yang$\Diamond$, and Thomas Huang\dag
 \\
\dag Beckman Institute, University of Illinois at Urbana-Champaign, Urbana, IL 61801, USA\\
\ddag Adobe Systems Inc, San Jose, CA 95110, USA\\
$\Diamond$ Snapchat Inc, Venice, CA 90291, USA\\
{\tt\small \{zwang119, yyang58, wang308, chang87, weihan3, t-huang1\}@illinois.edu} \\
{\tt\small  zhanwang@adobe.com} $\qquad$ {\tt\small jianchao.yang@snapchat.com}
}

\maketitle








\begin{abstract}
Deep learning has been successfully applied to image super resolution (SR). In this paper, we propose a deep joint super resolution (DJSR) model to exploit both external and self similarities for SR. A Stacked Denoising Convolutional Auto Encoder (SDCAE) is first pre-trained on external examples with proper data augmentations. It is then fine-tuned with multi-scale self examples from each input, where the reliability of self examples is explicitly taken into account. We also enhance the model performance by sub-model training and selection. The DJSR model is extensively evaluated and compared with state-of-the-arts, and show noticeable performance improvements both quantitatively and perceptually on a wide range of images.
\end{abstract}

\section{Introduction}

Super-resolution (SR) algorithms aim to construct a high-resolution (HR) image from one or multiple low-resolution (LR) inputs. Being ill-posed, SR has to resort to strong image priors, ranging from the simplest analytical smoothness assumptions, to more complicated statistical and structural priors \cite{Yang2012}, \cite{Fattal2010}. The most popular SR methods rely on a large and representative external set of image pairs to learn the mapping between LR and HR image patches \cite{Yang2010}. Those methods are known for their capabilities to produce plausible image appearances. However, there is no guarantee that an arbitrary input patch can be well matched or represented by a pre-chosen external set. When there are rarely matching features for the input, external examples are prone to produce either noise or oversmoothness \cite{Huang2010}. Meanwhile, image patches tend to recur within the same image \cite{Glasner2009}, \cite{Huang2010}, or across different image scales \cite{Fattal2010}. The self similarity property provides self examples that are highly relevant to the input, but only of a limited number. Due to the insufficiency of self examples, their mismatches often result in more visual artifacts. It is recognized that external and self example-based SR methods each suffer from their inherent drawbacks \cite{TIP}. 

The joint utilization of both external and self examples has been first studied for image denoising \cite{Zontak2011}. Mosseri et. al. \cite{Irani} proposed that image patches have different preferences towards either external or self examples for denoising. Such a preference is in essence the tradeoff between noise-fitting versus signal-fitting. Burger et. al. \cite{Harold} proposed a learning-based approach that automatically combines denoising results from an self example and an external example-based method. In SR literature, the authors in \cite{Dong} incorporated both a local autoregressive (AR) model and a nonlocal self similarity regularization term, into the sparse representation framework. Yang et. al. \cite{Yang2013} learned the approximated nonlinear SR mapping function from external examples with the help of in-place self similarity. More recently, a joint SR model was proposed in \cite{WACV}, \cite{TIP} \cite{wang2015designing}, to adaptively combine the advantages of both external and self examples. It is observed in \cite{TIP} that external examples contribute to visually pleasant SR results for relatively smooth regions, while self examples reproduce recurring singular features of the input. The complementary behavior has been similarly verified in the the image denoising literature \cite{Harold}.

More recently, inspired by the great success achieved by deep learning (DL) models in other computer vision tasks \cite{imagenet}, there is a growing interest in applying deep architectures to image SR. A Super-Resolution Convolutional Neural Network (SRCNN) was proposed in \cite{Tang}. Thanks to the end-to-end training and the large learning capacity of the CNN, the SRCNN obtains significant improvements over classical non-DL methods. In SRCNN, the information exploited for reconstruction is comparatively larger than that used in previous sparse coding approaches. However, the SRCNN has not taken any self similarity property into account. The authors in \cite{Shiguang} proposed the deep network cascade (DNC) to embed self example-based approach to auto-encoders (AEs). In each layer of the network, patchwise non-local self similarity search is first performed to enhance high-frequency details, and thus the whole model is not specifically designed to be an end-to-end solution. So far, there lacks a principled approach to utilize self similarity to regularize deep learning models, not only for SR, but also for general image restoration applications.

In this paper, we propose a unified deep learning framework, to joint utilize both the wealth of external examples, and the power of self examples specifically to the input. We name our proposed model \textit{deep joint super resolution} (\textbf{DJSR}). While the mutually reinforcing properties of external and self similarities are utilized in classical example-based methods \cite{Dong}, \cite{Yang2013}, \cite{TIP}, to our best knowledge, DJSR is the first to adapt deep models for joint similarities. The major contributions are summarized as multi-folds:
\begin{itemize}
\item We pre-train the model using an external set with data augmentations. We then fine-tune it using self-example pairs from the input image. Such a framework can be easily extended to other applications.
\item We propose to sample a large pool of self-example pairs using multi-scale self similarity, each of which is assigned a confidence weight during training. That alleviates the insufficiency of reliable self examples. 
\item We extend DJSR into several dedicated sub-models, and conduct selective training and patch processing. 
\end{itemize}
\textbf{Connecting SR to Domain Adaption} The idea of DJSR has certain connections to domain adaption or transfer learning \cite{sentiment}. For domain adaption, given a source domain having sufficient labeled data for training, and a target domain with insufficient labeled data and a different distribution, the problem is to have the model trained on the source domain generalize well on the target domain. In our setting, LR-HR pairs resemble the data-label tuples. The DJSR model is first learned on the source domain of external examples, and then adapted to the target domain of self samples from the testing image. That explains why DJSR could outperform previous models based on either external examples (applying source domain models directly to the target domain) or self examples (relying on target domain only to train models) from a domain adaption perspective.

\section{Pre-Training Using External Examples}


Several deep architectures have been explored for SR previously. The authors of DNC \cite{Shiguang} referred to a collaborative local auto-encoder (CLA) to be stacked to form a cascade. However, auto-encoders (AEs) rely mostly on fully-connected models and ignore the 2D image structure \cite{Vincent}. In SRCNN \cite{Tang}, a fully convolutional network is learned to predict the nonlinear LR-HR mapping. Such a model has a clear analogy to classical sparse coding methods \cite{Yang2012}.

In \cite{SCAE}, a Convolutional Auto Encoder (CAE) was proposed to learn features using a hierarchical unsupervised feature extractor while preserving spatial locality. CAEs can be stacked to form a Stacked Convolutional Auto Encoder (SCAE), where each layer receives its input from a latent representation of the layer below. It was further revealed in \cite{Vincent} that auto encoders are prone to learn trivial projections and the learned filters are usually subject to random corruptions. Denoising was thus suggested as a training criterion \cite{Vincent} to learn robust structural features, which also proves to be effective for image restoration tasks besides the original classification setting \cite{xie2012image}. We employ a Stacked Denoising Convolutional Auto-Encoder (SDCAE) \cite{SCAE} to reconstruct HR images from its stochastically corrupted LR versions. Such an architecture combines the intuitive idea of AEs, and the power of CNNs to capture 2-D structures efficiently. Note that more potential alternatives, such as SRCNN, can possibly be adapted here.

\begin{figure}[htbp]
\centering
\begin{minipage}{0.49\textwidth}
\centering {
\includegraphics[width=\textwidth]{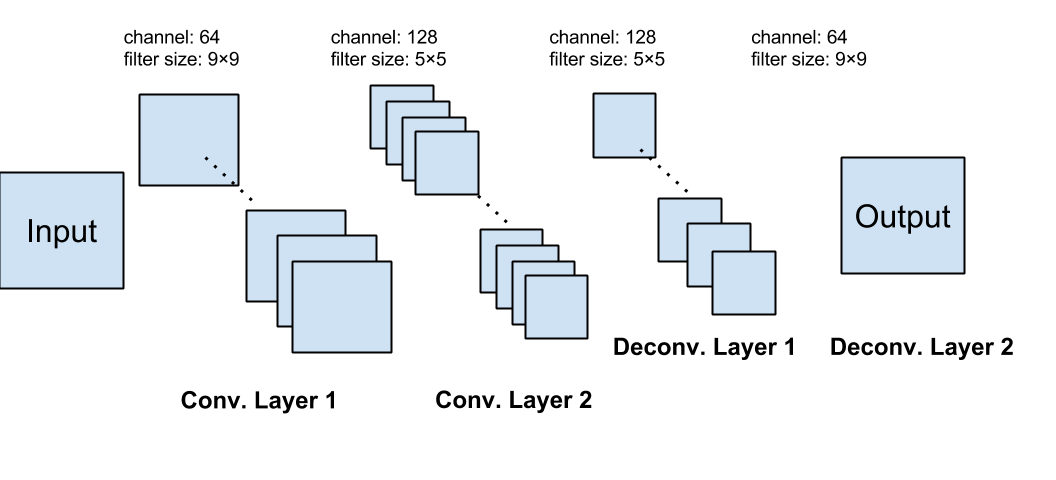}
}\end{minipage}
\caption{The SDCAE architecture for SR.}
\label{SDCAE}
\end{figure}

While there are multiple SCAE implementations available, we adopt the implementation by \cite{paine2014analysis} \footnote{https://github.com/ifp-uiuc/anna} as it has shown some improvements over the one in \cite{SCAE} on the CIFAR-10 benchmark. Their model is similar to the network in \cite{Deconv2010} but without using sparse coding, and introducing zero-biases \cite{memisevic2014zero} and ReLUs in the convolutional layers. To convert the SCAE into a SDCAE, all we need to do is to add a stochastic corruption (we use additive isotropic Gaussian noise) step operating on the input. Assuming that the original images are downsized by a scale $s$ to generate LR-HR example pairs for both training and evaluation, the SDCAE architecture is depicted in Fig. \ref{SDCAE}, where the input is a LR image and the output is its HR counterpart. The trained network fits SR with a factor of $s$.




\section{Fine-Tuning Using Self Examples}

\subsection{Self-example pairs by Multi-Scale Similarity}

In \cite{Shiguang}, the authors enhanced high-frequency by employing non-local self similarity (NLSS) search over the successive blurred and downscaled versions of the input image. By combining those internal matches, the estimated patch usually contains more abundant texture information. However, it overlooks the across-scale similarity properties of natural images \cite{Fattal2010}, \cite{Yang2013},  that singular features like edges and corners in small patches tend to repeat almost identically in their slightly upscaled versions. In addition, such a pre-processing step is not jointly optimized with the deep network cascade.

\begin{algorithm}[t]
\caption{Generate A Hierarchy of Self Examples Associated with Confidence Weights}
\begin{algorithmic}[1]
\REQUIRE Input image $\mathbf{Y}$, scaling factor $s$, number of scales $N$, number of self-example pairs per patch $m$.

\STATE Upsample $\mathbf{Y}$ for $\frac{N}{2}-1$ times with a factor of $s$.

\STATE Downsample $\mathbf{Y}$ for $\frac{N}{2}-1$ times with a factor of $s$.


\STATE For each patch in original $\mathbf{Y}$, find all spatially co-located patches in other $N-1$ resized versions.

\STATE For each co-located patch, find $\frac{m}{(N-1)}$ best matched examples from its immediate upscaled version, using the method in (\ref{freedman}). 

\STATE Make the co-located patch and each of its matched example a self-example pair.

\STATE Record the NN matching error for each match.

\STATE Take the negative exponents of all matching errors, and normalize them between [0,1]. Those will be used as the associated confidence weights.

\ENSURE $m$ self-example pairs per patch in $\mathbf{Y}$ with weights.
\end{algorithmic}
\end{algorithm}


Freedman and Fattal \cite{Fattal2010} applied the ``high frequency transfer'' method to search for the high-frequency component of a target HR patch, by NN patch matching across scales. Let $\mathbf{X}$ denote the HR image to be estimated from the LR input $\mathbf{Y}$. $\mathbf{X}_{ij}$ and $\mathbf{Y}_{ij}$ stand for the $(i,j)$-th ($i, j = 1, 2...$) patch from $\mathbf{X}$ and $\mathbf{Y}$, respectively.  Defining a linear interpolation operator $\mathcal{U}$ and a downsampling operator $\mathcal{D}$, for the input LR image $\mathbf{Y}$, we first obtain its initial upsampled image $\mathbf{X'} = \mathcal{U} (\mathbf{Y})$, and a smoothed input image $\mathbf{Y'} = \mathcal{D} (\mathcal{U} (\mathbf{Y}))$. Given the smoothed patch $\mathbf X'_{ij}$, the missing high-frequency band of each unknown patch $\mathbf X_{ij}$ is predicted by first solving a NN matching (\ref{freedman}):
\begin{equation}
\begin{array}{l}\label{freedman}
(m, n) = \arg\min_{(m, n) \in \mathcal{W}_{ij}} \| \mathbf Y'_{mn} - \mathbf X'_{ij}\|_F^2,
\end{array}
\end{equation}
where $\mathcal{W}_{ij}$ is defined as a searching window on image $\mathbf{Y'}$. With the co-located patch $\mathbf{Y}_{mn}$ from $\mathbf{Y}$, the high-frequency band $\mathbf{Y}_{mn} - \mathbf{Y'}_{mn}$ is pasted onto $\mathbf X'_{ij}$, i.e., $\mathbf X_{ij} = \mathbf X'_{ij}+ \mathbf{Y}_{mn} - \mathbf{Y'}_{mn}$.



Note the above methodology could be applied to construct self-example pairs $\{\mathbf{Y}_{ij}, \mathbf X_{ij}\}$ for a input image $\mathbf{Y}$. It is thus straightforward to consider adopting those self-example pairs to fine-tune our pre-trained network. However, two problems obstacle such a practice:
\begin{itemize}
\item \textbf{Insufficiency of Informative Examples} The amount of self examples is usually far less than the size of external training sets. For example, a $256 \times 256$ input  image can generate at most $(256 - 9 +1)^2$= 61,504 patches of a small size $9 \times 9$ (and thus the same amount of self-example pairs) and a minimum stride of 1 \cite{elad2006image}. Moreover, the information of those example pairs is also far less rich.
\item \textbf{Limited Reliability}  In essence, a part of input patches may be identified with few discernible repeating patterns. They might thus not be able to find good matches $\mathbf{Y'}_{mn}$ within the same image, which constitutes the visual artifacts in previous high frequency transfer methods \cite{Fattal2010}. Besides, The matching of $\mathbf{X'}_{ij}$ over $\mathbf{Y'}$ makes the core step of the high frequency transfer scheme. However, NN matching (\ref{freedman}) is not reliable under noise and outliers in LR images. 
\end{itemize}

To resolve the above raised concerns, we sample a hierarchy of self examples from multiple scales, each of which is associated with a confidence weight calculated by the NN matching error from (\ref{freedman}). The key idea is to exploit cross-scale patch redundancy embedded between multiple neighborhood scales. The steps are outlined in Algorithm I. 

\subsection{Weighted Back Propagation}
The self-example pairs obtained from Algorithm 1 can be used to fine-tune the pre-trained SDCAE, making it specially adapted for the input. To incorporate the reliability of the self-example pairs into the process, a variant of standard back propagation, called \textit{Weighted Back Propagation} (\textbf{WBP}), is developed to alleviate the negative impacts of bad examples, without sacrificing the benefits of abundant training data. In particular, assuming that $\omega$ is the normalized confidence weight for the current self-example pair, let $\eta_f$ denote the learning rate for fine tuning and $\delta$ the gradient, the weight matrices $W_i$ ($i$ is the layer index) are updated as:
\begin{equation}
\begin{array}{l}\label{WBP}
\delta_{i+1}= \omega \cdot (\eta_f \cdot \frac{\partial{L}}{\partial{W_i}} + 0.9 \cdot \delta_i), \quad W_{i+1} = W_i + \delta_{i+1}
\end{array}
\end{equation}
Note each example pair possesses a different (and pre-calculated) $\omega$. Such an importance weighting strategy has been commonly applied to transfer learning problems \cite{liu2014robust}. Yet to our best knowledge, there is no similar work in deep learning. In all experiments, the learning rate $\eta_f$ for fine-tuning is set to be $0.5$ by default, and will not be annealed. That large value is empirically found to work well, leading to a better presence of self similarity in final SR results.

%
%

\section{Sub-model Training and Selection}

Previous DL-based image SR methods aim at learning one model that is capable to represent various image structures. Such a model lacks the adaptivity to local structures. In some cases, It might also lead to a model of overly high complexity and redundancy \cite{Dong}. When learning regressors from LR to HR patches, it is observed that regressors have different specialities at dealing with certain patches \cite{daijointly}. Following this idea, external examples are partitioned into many clusters, each of which consists of patches with similar patterns and can be used to pre-train a sub SDCAE model.  Next, for each input patch, the most relevant sub-model is first selected and then fine-tuned by its own self-example pairs. Since a given patch can be better represented by the adaptively selected sub-models, the whole HR image can be more accurately reconstructed.

Provided with an external set, we first use the high-pass filtering output of each LR patch as the feature for clustering. It allows us to focus on the edges and structures of image patches. We then adopt K-means algorithm to partition the whole set into $K$ clusters, where $\mu_i$ denotes the centroid of $i$-th cluster, $i=1,2,..., K$. During model fine-tuning (and testing), the best sub-model is chosen based on the minimum distance between the LR patch and the centroids. As suggested by \cite{Dong}, let $U$ = [$\mu_1$, $\mu_2$, ...$\mu_K$], its PCA transformation matrix is obtained by applying SVD to the co-variance matrix of $U$. We can compute the distance between input patch and cluster centroids more robustly in the subspace spanned by the most significant eigenvectors \cite{Dong}. 


%
%
%

\section{Experiments}

\subsection{Implementation}

The SDCAE is learned from an external training set with 91 images \cite{Yang2010}, which is also adopted in \cite{Tang}. The 91-image dataset can be decomposed into 24,800 sub-images of size 33 $\times$ 33 for training purpose, which are extracted from original images with a stride of 14. For each LR patch, we subtract its mean and normalize its magnitude, which are later put back to the recovered HR patch. While data augmentation is not adopted in SRCNN, we believe that it plays an important role in training DJSR, to help it focus on meaningful visual features rather than artifacts in training images. We add the following distortions to training images:
\begin{itemize}
\item \textit{Translation:} random x-y shifts between [-4, 4] pixels.
\item \textit{Rotation:} affine transform with random parameters.
\item \textit{Zoom:} random scaling factors between [1/1.2, 1.2]. Note it has to keep the ratio $s$ unchanged.
\end{itemize}
SDCAE can also be viewed as a data augmentation way by adding noise. We train SDCAE on sub-images, using stochastic gradient descent with a constant momentum of 0.9, and a learning rate $\eta_p$ of 0.01 (we do not anneal it through training). Mean Squared Error (MSE) is used as the loss function. 

Since cross-scale self similarity performs best at small scales \cite{Fattal2010}, we stick to a small upscaling factor $s$ (1.2 by default) for model training, unless otherwise specified. To achieve any targeted upscaling factor $s_t$, we zoom up an image repeatedly using the learned DJSR model  until it is at least as large as the desired size. Then a bicubic interpolation is used to downscale it to the target resolution if necessary. We do not conduct extra joint optimization on the resulting network cascade. The proposed networks are implemented using the CUDA ConvNet package \cite{imagenet} and the ANNA open source library \cite{paine2014analysis}, and run on a workstation with 12 Intel Xeon 2.67GHz CPUs and 1 GTX680 GPU.

For color images, we apply SR algorithms to the illuminance channel only, while interpolating the color layers (Cb, Cr) using plain bi-cubic interpolation. However our model is flexible to process more color channels without altering the network design. To avoid border effects, all the convolutional layers have no padding, and the network produces a smaller central output \cite{Tang}.

\subsection{Model Analysis}

\begin{figure}[htbp]
\centering
\begin{minipage}{0.50\textwidth}
\centering \subfigure[] {
\includegraphics[width=\textwidth]{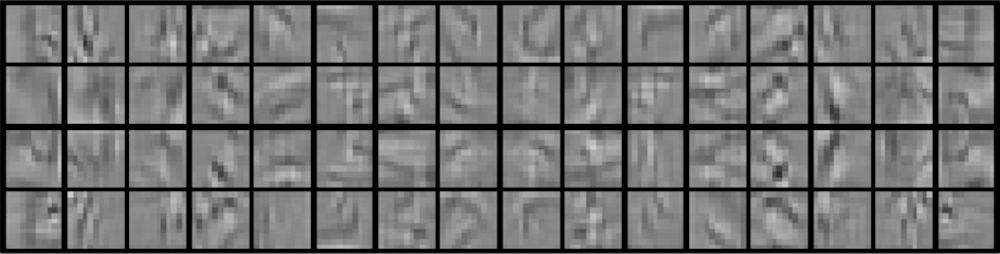}
}\end{minipage}
\begin{minipage}{0.50\textwidth}
\centering \subfigure [] {
\includegraphics[width=\textwidth]{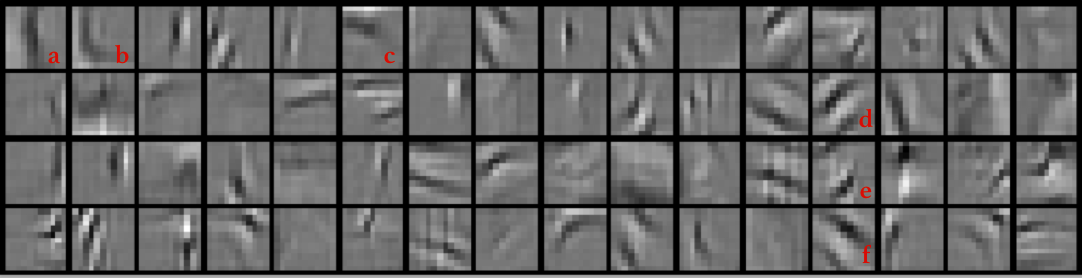}
}\end{minipage}
\begin{minipage}{0.50\textwidth}
\centering \subfigure [] {
\includegraphics[width=\textwidth]{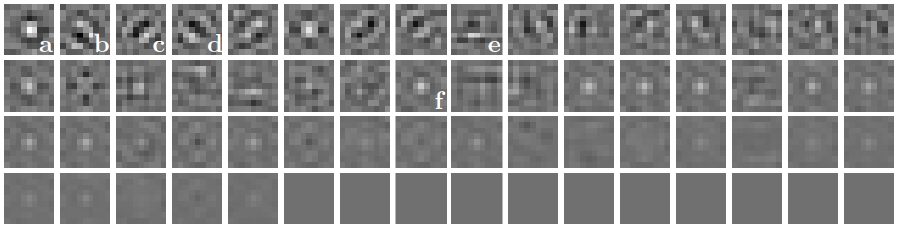}
}\end{minipage}
\caption{(a) the first layer convolutional filters learnt by SDCAE without any data augmentation; (b)  the first layer convolutional filters learnt by SDCAE with data augmentations; (c) the first layer convolutional filters in SRCNN (from the original Fig. 5 in \cite{Tang}). }
\label{filters}
\end{figure}


%
%

\noindent \textbf{Validating Pre-Training} We visualize the learned convolutional filters in the first layers of SDCAEs without and with augmentations. They are trained with a relatively large scaling factor $s$ = 2, so as to be compared with the first-layer filter visualizations of SRCNN, as depicted in Fig. \ref{filters}. The training process takes around 7 hours.  Both SDCAEs and SRCNN have 64 channels of $9 \times 9$ convolutional filters in the first layer. While there is hardly any recognizable structural features from the filters in (a), the introduction of data augmentations leads to much more clear and interpretable filter responses in (b), from simple edge (curve) detectors at different directions (e.g., a, b and c), to more sophisticated texture descriptors (e.g., d, e and f). On the other hand, since the first layer of SRCNN is designed for patch extraction and representation, it is natural that its learned filters show different from ours. One interesting observation is that SRCNN suffers from several ``dead'' filters, whose weights are all nearly zeros (as discussed in  \cite{Tang}), whereas almost all filters of SDCAE are fairly strong and diverse. Further, the SDCAE with augmentations obtains an average PSNR of 36.43 dB when testing on the Set 5 \cite{Tang} (with no fine-tuning applied), where we see a notable performance improvement of 1.44dB compared to the case without augmentations (35.01 dB).

\noindent \textbf{Understanding Fine-Tuning} During fine-tuning, we sample LR patches from the input $\mathbf{Y}_{ij}$ with a default size of $15 \times 15$ and stride of 1. In this section, we take the \textit{Baby} LR image of size $256 \times 256$ for example,which will result in 58,564 patches. Assuming SDCAE has been pre-trained, by default, we fix $N$ = 5 (defined in Algorithm 1), which means the hierarchy will contain 2 upscaled layers (by factors of 1.2 and 1.44) and two downscaled layers (by factors of 1/1.2 and 1/1.44). Fine-tuning a trained SDCAE on \textit{Baby} takes less than 1 hour, and it could be potentially accelerated to a large extent by avoiding working on those homogeneous regions \cite{Yang2013}. We will then investigate the influences of the parameter $m$ (that controls the amount of self-example pairs), and the effects of tuning the learning rate $\eta_f$, as well as the effects of WBP algorithm.

\begin{figure}[htbp]
\centering
\begin{minipage}{0.40\textwidth}
\centering {
\includegraphics[width=\textwidth]{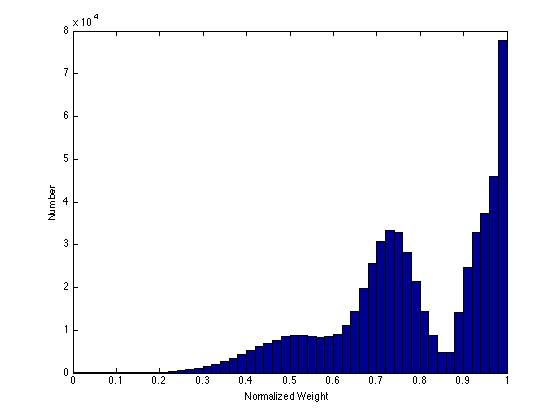}
}\end{minipage}
\caption{The histogram of normalized weight values of all self-example pairs obtained on \textit{Baby} ($m$=8).}
\label{peak}
\end{figure}

 \begin{table}[h]
 \begin{center}
 \caption{The effects of $m$ on the average effective volume and the final PSNR (dB) after fine-tuning, when upscaling \textit{Baby} image for 2 times (PSNR is 37.91dB before fine-tuning).}
 \label{volume}
 \vspace{0.1em}
 \begin{tabular}{|c|c|c|c|c|c|}
 \hline
$m$ &  4 &  8  & 12  & 16  \\
 \hline
$\frac{m}{N-1}$ & 1 & 2  & 3 & 4  \\
  \hline
 $V$ & 234,256   &   468,512    &  702,768   &   937,024    \\
  \hline
$V_e$ &  178, 597 &  365,251 &  397,242 & 400,272  \\
 \hline
 $V_e^a$ & 0.7624 & 0.7796 & 0.5653   & 0.4272  \\
 \hline
PNSR&  38.01 & 38.87  &  38.90  & 38.91  \\
 \hline
$\text{PNSR}_{NW}$&  38.03 & 38.41  &  38.00  & 37.51  \\
 \hline
 \end{tabular}
 \end{center}
 \end{table}

\begin{figure*}[htbp]
\centering
\begin{minipage}{0.33\textwidth}
\centering \subfigure [SRCNN] {
\includegraphics[width=\textwidth]{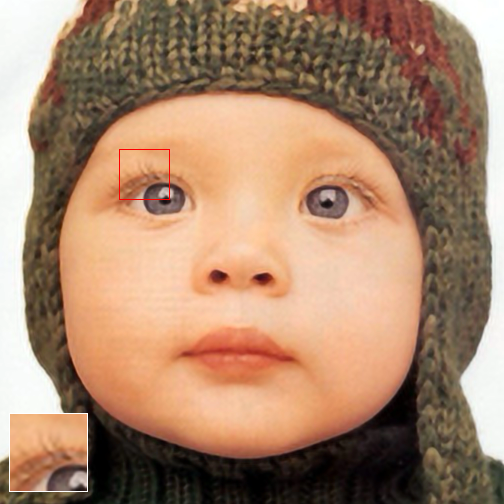}
}\end{minipage}
\begin{minipage}{0.33\textwidth}
\centering \subfigure [DNC] {
\includegraphics[width=\textwidth]{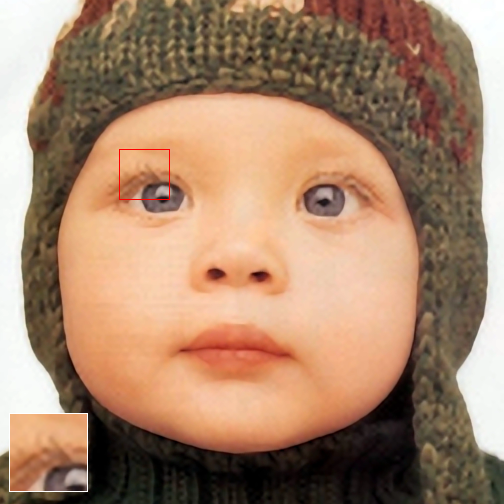}
}\end{minipage}
\begin{minipage}{0.33\textwidth}
\centering \subfigure [DJSR] {
\includegraphics[width=\textwidth]{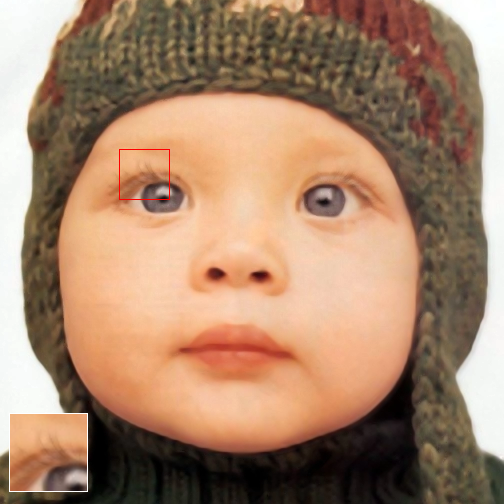}
}\end{minipage}
\caption{3$\times$ SR results of the \textit{Baby} image by: (a) SRCNN, PSNR = 29.22 dB, SSIM = 0.9047; (b) DNC,  PSNR = 26.65 dB, SSIM = 0.8490; (c) DJSR, PSNR = 28.74 dB, SSIM = 0.9074.}
\label{SR1}
\end{figure*}


   \begin{table}[t]
 \begin{center}
 \caption{The effects of $\eta_f$ on the final PSNR (dB) after fine-tuning, when upscaling \textit{Baby} image for 2 times (The PSNR is 37.91dB before fine-tuning).}
 \label{rate}
 \vspace{0.1em}
 \begin{tabular}{|c|c|c|c|c|c|c|c|}
 \hline
$\eta_f$  &  0.01 &   0.1  & 0.3 & 0.5 & 0.6 & 0.8  \\
 \hline
PNSR&  37.91 &  38.12 & 38.44 & 38.87 & 38.36 & 37.99 \\
 \hline
 \end{tabular}
 \end{center}
 \end{table}
 

Fig. \ref{peak} depicts the distribution of normalized weight values of all self-example pairs obtained on \textit{Baby}, with $m$ = 8. Notably, two peaks appear on the histogram, one largest peak near the weight value of 1, and the other lower one is centered around 0.75. Further observations reveal that the first peak corresponds to those in-place self examples as in \cite{Yang2013}, which follow the local scale invariance property and are usually very accurate matches. The second peak, with relatively larger errors, mostly corresponds to the non-local similar examples \cite{Fattal2010}. 


To further understand how the amount of self examples and their weights influence the fine-tuning, we introduce several measurements: Let $V$ denote the \textit{total volume} of self-example pairs (thus $V = 58, 564 \times m$). Define the \textit{effective volume} $V_e$ as the (rounded) summations of all normalized weights, and the \textit{average effective volume} $V_e^a = \frac{V_e}{V}$. $V_e^a$ can be viewed as an indicator on how reliable and representative the chosen self-example set is. As shown in Table \ref{volume}, with $m$ growing from 4 to 8, both $V_e$ and $V_e^a$ increase, implying that self similarity is better exploited. The PSNR improvement after fine-tuning also becomes more substantial. However, when $m$ continues going up, both $V_e$ and PSNR reach the plateau, whereas $V_e^a$ decreases dramatically. That clearly manifests that little self similarity information remains to be excavated, and the self example sets assumably turn redundant. Therefore, $m$ is set as 8 by default hereinafter. The last row of Table \ref{volume} lists the PSNR results obtained from fine-tuning with standard back propagation, denoted as $\text{PNSR}_{NW}$. It is noteworthy that without taking the confidence weights into consideration, more self-example pairs may even harm the SR performance of the pre-trained model when $V_e^a$ drops. The WBP algorithm shows quite robust when more self examples are used in fine-tuning. 


Table \ref{rate} examines the PNSR changes with varying $\eta_f$. We observe that a small learning rate (such as 0.01) does not bring any notable improvement to final results. Until $\eta_f$ = 0.5 (the empirical default value), a growing $\eta_f$ leads to a monotone increase in final PSNR results. That is interpretable, as self examples in any way do not contain as sufficient and diverse information as external examples do; a large learning rate can strengthen their influences on the pre-trained model. It may also help overcome some local minima. However, when $\eta_f$ is further improved beyond 0.5, the performances start to be undermined, and more fluctuations are observed during the  convergence process ($\eta_f$ = 0.8 actually does not lead to a stable convergence). 

\begin{figure*}[htbp]
\centering
\begin{minipage}{0.33\textwidth}
\centering \subfigure [SRCNN] {
\includegraphics[width=\textwidth]{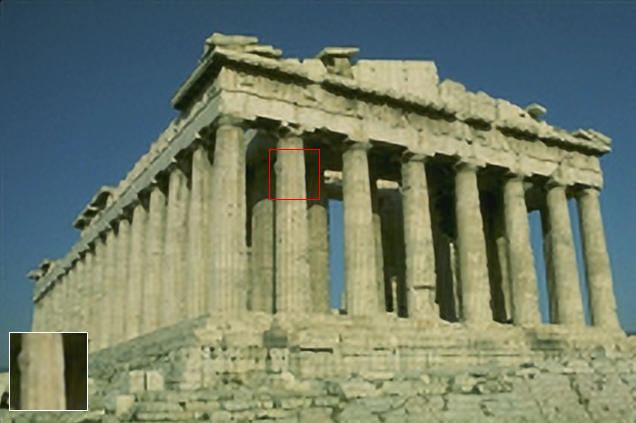}
}\end{minipage}
\begin{minipage}{0.33\textwidth}
\centering \subfigure [DNC] {
\includegraphics[width=\textwidth]{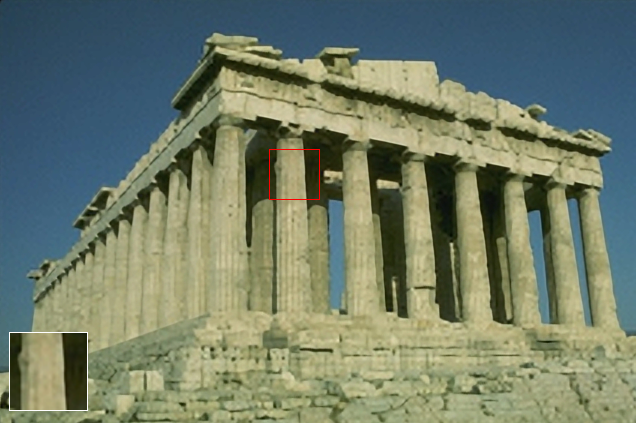}
}\end{minipage}
\begin{minipage}{0.33\textwidth}
\centering \subfigure [DJSR] {
\includegraphics[width=\textwidth]{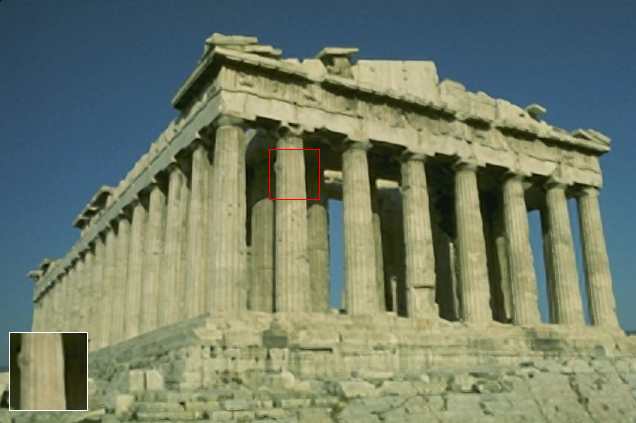}
}\end{minipage}
\caption{3$\times$ SR results of the \textit{Roman} image by: (a) SRCNN, PSNR = 29.97 dB, SSIM = 0.9250; (b) DNC,  PSNR = 30.08 dB, SSIM = 0.9293; (c) DJSR, PSNR = 30.69 dB, SSIM = 0.9337.}
\label{SR2}
\end{figure*}

\begin{figure*}[htbp]
\centering
\begin{minipage}{0.33\textwidth}
\centering \subfigure [SRCNN] {
\includegraphics[width=\textwidth]{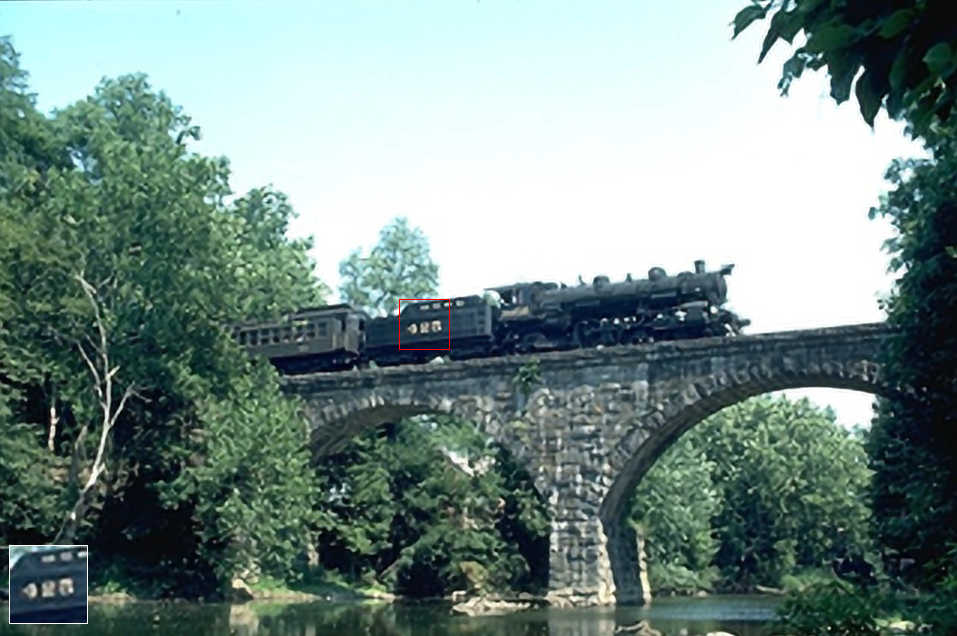}
}\end{minipage}
\begin{minipage}{0.33\textwidth}
\centering \subfigure [DNC] {
\includegraphics[width=\textwidth]{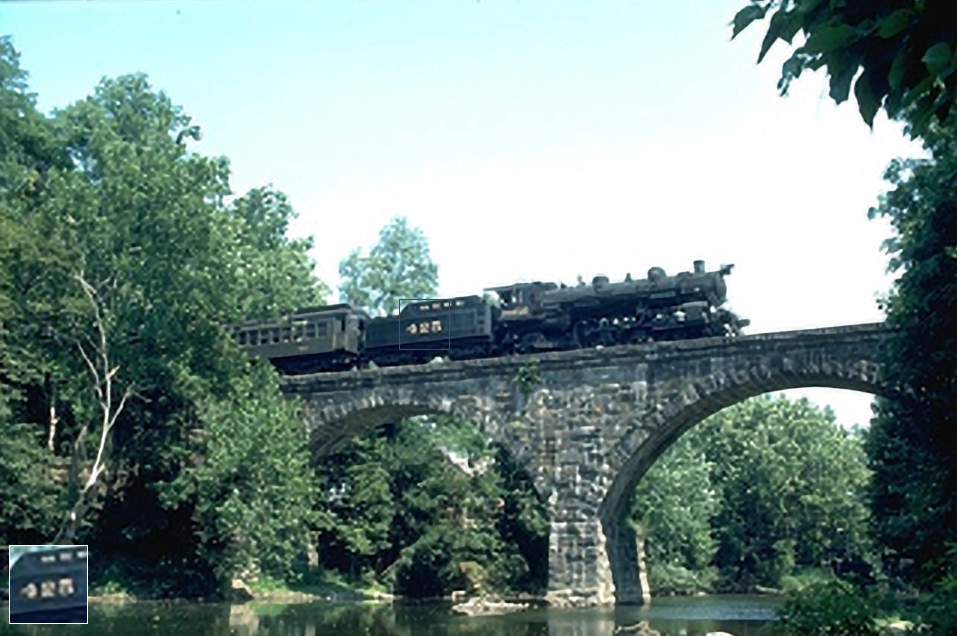}
}\end{minipage}
\begin{minipage}{0.33\textwidth}
\centering \subfigure [DJSR] {
\includegraphics[width=\textwidth]{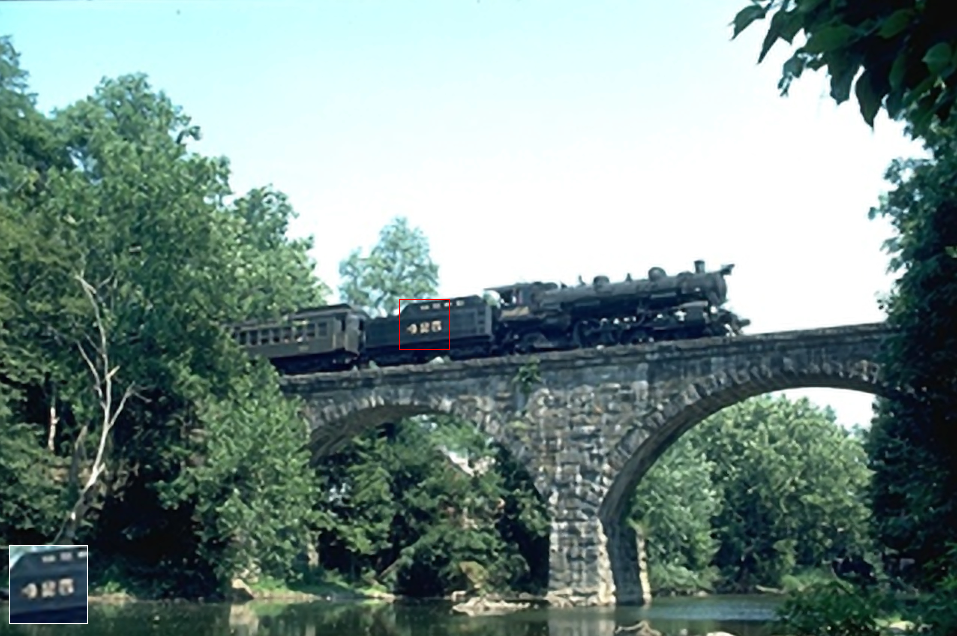}
}\end{minipage}
\caption{3$\times$ SR results of the \textit{Train} image by: (a) SRCNN, PSNR = 29.67 dB, SSIM = 0.9614; (b) DNC, PSNR = 28.02 dB, SSIM = 0.9392; (c) DJSR, PSNR = 30.57 dB, SSIM = 0.9706.}
\label{SR3}
\end{figure*}


\subsection{Comparison with State-of-the-Arts}

We first compare DJSR qualitatively with two recent DL-based SR methods, SRCNN \cite{Tang} \footnote{Results by using the original implementation available at: http://mmlab.ie.cuhk.edu.hk/projects/SRCNN.html} and DNC \cite{Shiguang} \footnote{Results provided by the authors: http://vipl.ict.ac.cn/paperpage/DNC}. Fig. \ref{SR1}, \ref{SR2}, and \ref{SR3} demonstrate visual comparisons on three natural images, \textit{Baby}, \textit{Roman}, and \textit{Train}, respectively; all are upsampled by a factor $s_t$ of 3. The zoomed regions are also displayed. SCRNN performs reasonably well on \textit{Baby} and \textit{Train} images, but are visually worse than DNC on the \textit{Roman} image, since \textit{Roman} are abundant in repeating textures on the pillars of the Parthenon, making self similarity especially powerful. DJSR produces image patterns with shaper boundaries and richer textures (see the zoomed pillar regions on \textit{Roman}, and the numbers on \textit{Train}), and suppresses the jaggy and blockiness artifacts discernibly better. 

PSNR and SSIM \cite{SSIM} are used to evaluate the performances quantitatively (only the luminance channel is considered).  While all three deep networks are optimized under a MSE (equivalent to PSNR) loss, DJSR is slightly worse than SRCNN on \textit{Baby} in terms of PSNR, but obtains the best performances on both \textit{Roman} and \textit{Train} images. What is more, we notice that DJSR is particularly more favorable by SSIM, which measures image quality more consistently with human perception than PSNR. The observation is further verified on the commonly-adopted Set 5 and Set 14 datasets. Such an advantage can be owed to our fine-tuning step, which further enhances the generic model by exploiting the self similar structures of the input. Table \ref{set} compares the average PSNR and SSIM results of the DJSR and SRCNN \footnote{DNC is not included, since neither the original codes nor any reported result on the two sets are unavailable. A part of data in Table \ref{set} is from \cite{Tang}}, as well as a few other classical non-DL SR methods, on the Set 5 and Set 14 datasets. DJSR obtains an overall competitive performance, and especially gains a consistent advantage over others in SSIM.

\subsection{Evaluation of Sub-models}

\begin{figure}[htbp]
\centering
\begin{minipage}{0.38\textwidth}
\centering {
\includegraphics[width=\textwidth]{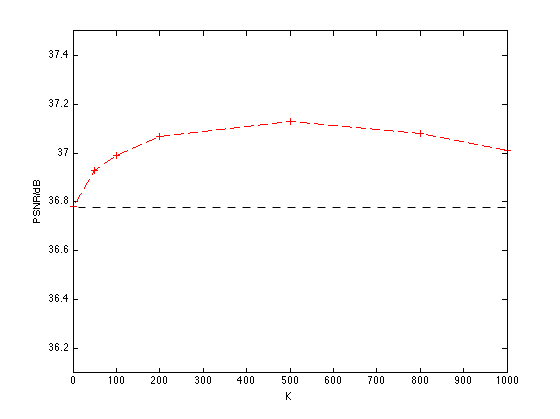}
}\end{minipage}
\caption{The average PSNR results on Set 5 for upscaling by 2 times, with varying $K$.}
\label{cluster}
\end{figure}

While the DJSR model could very well compete against the state-of-the-arts, there is still potential room for improvements, by training a group of sub-models and selecting the optimal sub-models for each patch. The number of clusters $K$ is a parameter to be pre-determined. Specifically, we train the sub-models under different $K$ values, and applied them to upscale LR images in Set 5 ($s_t$ = 2). Fig. \ref{cluster} records the average PSNR values (the black dash line denotes the original DJSR model, i.e., $K$=1). We can see that when $K$ increases from 50 to 500, the PSNR results gradually raise, as each sub-model is developed to describe a smaller subset of similar image patches more precisely. Yet a slight performance drop occurs at $K$ =800 and continues when $K$ increases to 1,000. A closer look into the clusters demonstrates that when $K$ becomes too large, many clusters contain only as few as thousands of examples, which are inadequate for training a deep network. Such "chaos" sub-models will finally hamper the overall performance.

\begin{table*}[t]
\begin{center}
\caption{Average PSNR (dB) and SSIM performances comparisons on the Set 5 and Set 14 datasets}
\label{set}
\vspace{0.5em}
\begin{tabular}{|c|c|c|c|c|c|c|c|}
\hline
 & & Bicubic & Sparse Coding \cite{Yang2010} & Freedman et.al. \cite{Fattal2010}& A+ \cite{A+} & SRCNN \cite{Tang} &  DJSR \\
\hline
$\multirow{2}{*}{\textit{Set 5, $s_t$=2}}$ &PSNR  & 33.66 & 35.27 & 33.61 & 36.24 & 36.66 & \textbf{36.78}\\
\cline{2-8}
$$ &SSIM & 0.9299 & 0.9540 & 0.9375 & 0.9544 & 0.9542 & \textbf{0.9550} \\
\hline
$\multirow{2}{*}{\textit{Set 5, $s_t$=3}}$ &PSNR & 30.39 & 31.42 & 30.77 & 32.59 & \textbf{32.75} & 32.65\\
\cline{2-8}
$$ &SSIM  & 0.8682 & 0.8821 & 0.8774 &0.9088 & 0.9090 & \textbf{0.9161} \\
\hline
$\multirow{2}{*}{\textit{Set 14, $s_t$=2}}$ &PSNR & 30.23 & 31.34 & 31.99 & \textbf{32.58} & 32.45& 32.51  \\
\cline{2-8}
$$ &SSIM & 0.8687 & 0.8928 & 0.8921 & 0.9056 & 0.9067 & \textbf{0.9097}  \\
\hline
$\multirow{2}{*}{\textit{Set 14, $s_t$=3}}$ &PSNR & 27.54 & 28.31 & 28.26 & 29.13 & 29.60 & \textbf{29.96}  \\
\cline{2-8}
$$ &SSIM  & 0.7736 & 0.7954 & 0.8043 & 0.8188 & 0.8215 & \textbf{0.8229} \\
\hline
\end{tabular}
\end{center}
\end{table*}

Fig. \ref{cluster} reminds us that while dividing sub-models is usually supposed to be helpful, an improper choice of $K$ can impact the results negatively. The optimal selection of $K$ is a nontrivial task, and is subject to the bias and variance tradeoff \cite{Dong}. If $K$ is too small, the boundaries between clusters will be smoothed out and the distinctiveness of sub-models are compromised. On the other hand, an overly larger $K$  will make a part of sub-models unreliable. A simple heuristics is adopted in our experiments: the training dataset is first partitioned into 100 clusters; next, those ÒfragmentÓ small clusters (e.g., containing less than 500 samples) are merged into their neighboring clusters. That will usually lead to $K$ between [40, 50]. We find that strategy leads to a good and stable performance improvement, after we try it on several different training sets.

\section{Conclusion}

In this paper, we investigate a deep joint super resolution (DJSR) model, to exploit external and self similarities for image SR in a unified framework.  We utilize external examples to pre-train the model, and fine-tune it using sufficient self examples weighted by their reliability. We thoroughly analyze the model and interpret its behaviors. DJSR is compared with several state-of-the-art SR methods (both DL and non-DL) in our experiments, and shows a visible performance advantage both quantitatively and perceptually. Similar approaches can be extended to many other image restoration applications.

%

%
{\small
\bibliographystyle{ieee}
\bibliography{djsr}
}


\end{document}